%% file: joyinside.tex
\documentclass[10pt,letterpaper]{jingdong}
\usepackage{amsmath,amssymb,amsfonts}
\usepackage{enumitem}
\usepackage{tikz}
\input{authors.tex}

\newcommand{\previewincludegraphics}[2][]{%
  \IfFileExists{#2}{\includegraphics[#1]{#2}}{%
    \PackageWarning{joyinside-preview}{Missing figure: #2}%
    \fbox{\parbox[c][32mm][c]{0.90\linewidth}{%
      \centering\normalfont\small Figure file not present locally\par
      \texttt{\detokenize{#2}}}}%
  }%
}

\title{JoyInside-VLM-Agent: Product-Conditioned Multimodal Policy Learning for Embodied Agents}
\abstract{\input{abstract.tex}}

\begin{document}
\maketitle

\section{Introduction}

\begin{figure}[t!]
  \centering
  \previewincludegraphics[width=\linewidth]{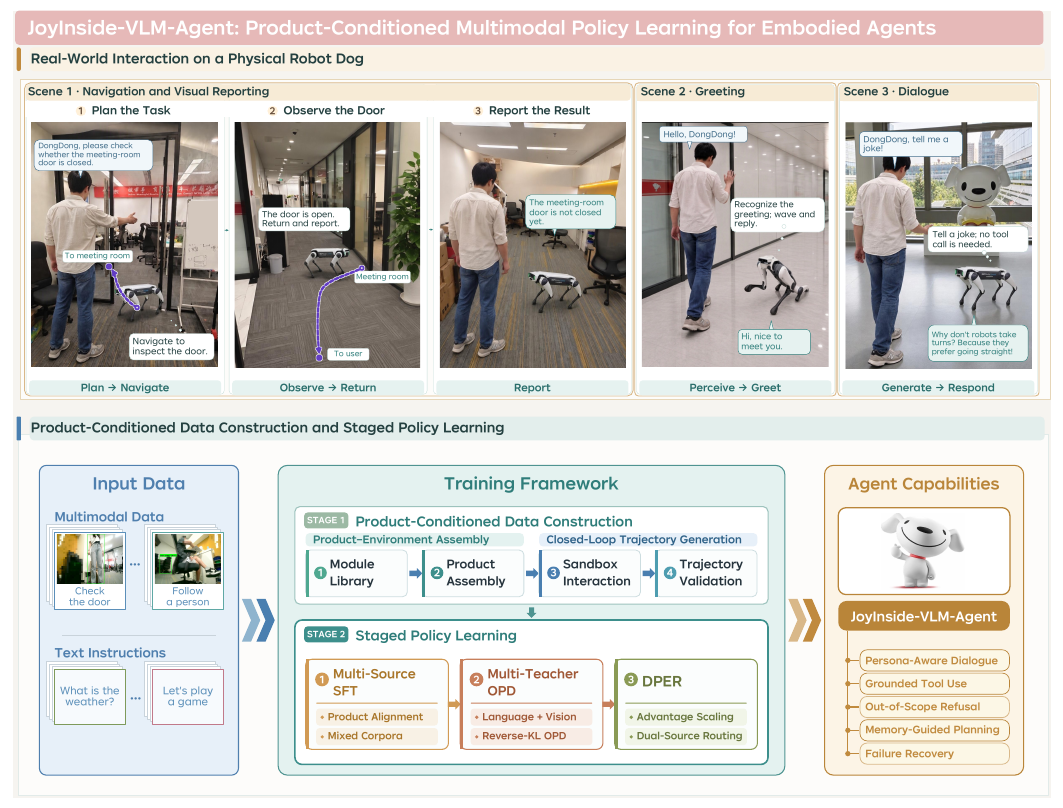}
  \caption{\textbf{JoyInside-VLM-Agent overview.} \textit{Top:} Real-world quadruped interactions show navigation with visual reporting (left), greeting (center), and dialogue (right). \textit{Bottom:} Product-conditioned data construction (Stage 1) supplies trajectories and executable environments for SFT, multi-teacher OPD, and DPER (Stage 2). The right panel summarizes agent capabilities.}
  \label{fig:teaser}
\end{figure}

\noindent\hspace*{\parindent}Embodied agents deployed on robotic platforms and smart hardware must translate multimodal observations and user instructions into executable behavior. Across products ranging from bipedal, quadruped, and wheeled robots to AI-enabled plush toys, the appropriate response to a given request depends on the agent's role, capabilities, and operating constraints. Changes in user goals, new visual evidence, and execution failures can further invalidate earlier plans. We study how to train a shared multimodal policy that adapts to product specifications and revises its decisions as interactions unfold. Figure~\ref{fig:teaser} presents the training framework and deployment examples.

Multimodal training has demonstrated positive transfer across tasks and robotic embodiments \cite{driess2023palme,oxe2023}, while context-aware systems use user intent and environmental feedback to guide interaction \cite{king2024sasha,chang2024worldscribe}. However, reported failures include actions beyond device capabilities and descriptions that no longer match the user's current scene or intent \cite{king2024sasha,chang2024worldscribe}. These observations motivate training policies that combine transferable capabilities with adherence to product constraints and sustained alignment with the user's current goal.

This objective imposes two requirements on training. The data must cover diverse product configurations, goal revisions, and failure recovery while preserving the correspondence between actions and observations. The learning signal must reflect reasoning quality, compliance with product constraints, and progress toward the current goal. Trajectory-level rewards summarize these criteria but do not specify how to distribute credit across tokens. Using additional evidence during training therefore requires selecting an appropriate source for each token and controlling how that evidence affects policy updates.

We present JoyInside-VLM-Agent, a multimodal agent developed through product-conditioned data construction and staged policy learning. Our pipeline combines product roles, workflows, and constraints with available tools and skills to build executable environments. A closed-loop sandbox generates multimodal interaction trajectories, which undergo execution checks and semantic validation. Multi-source supervised fine-tuning establishes behavior consistent with product specifications, while multi-expert on-policy distillation integrates complementary language and visual capabilities. Reinforcement learning then refines the policy using interaction-specific rewards.

To guide reinforcement learning, we combine process-quality scores from a Generative Process Reward Model (GPRM), multi-judge rankings of the same candidate trajectories, and task-queue assessments of goal alignment and cross-turn continuity. For token-level policy optimization, we introduce DPER (Dual-Source Privileged On-Policy Self-Distillation with Token-Level Evidence Routing), building on RLSD's sign-preserving advantage modulation \cite{yang2026selfdistilledrlvr} and DOPD's dual-source routing \cite{yu2026dopd}. A frozen privileged teacher and a privileged student sharing the rollout policy's parameters score tokens along the same student-generated trajectory. DPER selects evidence based on their disagreement and support for each sampled token. The selected evidence rescales token-level advantages while preserving their reward-derived signs. Privileged information is used only during training.

We evaluate JoyInside-VLM-Agent on public tool-use benchmarks and the proprietary JoyInside-Devices and JoyInside-Robotics datasets. The model achieves a BFCL v4 overall score of 77.44, outperforming the evaluated baselines. Case studies on a physical quadruped robot provide qualitative evidence of visually grounded search, replanning in response to execution feedback, and honest failure reporting.

Our contributions are as follows:
\begin{itemize}[leftmargin=1.5em,label=\textbullet]
  \item \textbf{Product-conditioned trajectory construction.} We compose executable product configurations and generate verified multimodal trajectories covering product variation, goal revisions, and failure recovery.

  \item \textbf{DPER for policy optimization.} We route token-level evidence between two privileged sources to scale token advantages while preserving their reward-derived signs.

  \item \textbf{Multi-source interaction rewards.} We combine process evaluation, multi-model rankings, and task-queue consistency into a learning signal for reasoning quality, adherence to product constraints, and alignment with evolving goals.

  \item \textbf{JoyInside-VLM-Agent and its evaluation.} We integrate these components into a unified training framework and assess the resulting model through public benchmarks and proprietary evaluation sets and qualitative quadruped case studies.
\end{itemize}

\section{Related Work}

\subsection{Embodied and Context-Aware Interaction}

\noindent\hspace*{\parindent}PaLM-E learns shared representations for robotic and vision-language tasks \cite{driess2023palme}, while Open X-Embodiment studies cross-platform transfer using pooled interaction data \cite{oxe2023}. Context-aware assistants address how observations and user intent guide behavior: Sasha grounds smart-home requests in available devices and user feedback \cite{king2024sasha}; GazePointAR combines gaze, pointing, and dialogue history to resolve references \cite{lee2024gazepointar}; and WorldScribe adapts visual descriptions to intent and environmental changes \cite{chang2024worldscribe}. Together, these lines connect shared capabilities with decisions grounded in the current interaction.

Interactive benchmarks further make behavioral constraints explicit. $\tau$-bench evaluates conversations under domain policies \cite{yao2024taubench}, while $\tau^2$-Bench considers user--agent coordination in a shared environment \cite{barres2025tau2bench}. WildToolBench covers intentions distributed across turns and transitions among tool use, conversation, and clarification \cite{yu2026wildtoolbench}. These settings motivate evaluating whether behavior respects constraints and follows the current goal. Our work addresses the corresponding training problem across robotic platforms and smart hardware, conditioning a shared policy on each product's role, capabilities, constraints, and interaction state.

\subsection{Multimodal Trajectory Construction}

\noindent\hspace*{\parindent}Synthetic data methods expand agent training beyond available interaction logs. ToolACE constructs diverse function-calling dialogues with rule- and model-based verification \cite{liu2024toolace}, and T3-Agent generates and verifies multimodal queries and tool-use trajectories \cite{gao2025mmagent}. Their focus on coverage and validity provides a basis for learning tool interaction. Training for dynamic tasks additionally requires representing the consequences of intermediate actions and the decisions that follow unexpected feedback.

Agent-R constructs revision trajectories through search and model-guided critique, connecting erroneous behavior to successful continuations \cite{yuan2025agentr}. Agent World Model builds executable, stateful environments for interaction and reinforcement learning \cite{wang2026agentworldmodel}. These approaches expose learners to recovery behavior or environmental transitions. Our pipeline couples such execution feedback with compositional product specifications and visual observations, covering both recovery toward an existing goal and adaptation after the user changes it. We jointly vary product roles, workflows, constraints, and capabilities within executable environments.

\subsection{Post-Training and Reward-Guided Optimization}

\noindent\hspace*{\parindent}AgentTuning and Agent-FLAN study how instruction tuning develops agent behavior from demonstrations \cite{zeng2024agenttuning,chen2024agentflan}. Generalized knowledge distillation obtains teacher feedback on student-generated sequences, addressing the mismatch between fixed demonstrations and the learner's behavior \cite{agarwal2024gkd}. H-OPD combines vision-language and text-only teachers through visual-description transfer and token-level arbitration \cite{yin2026hopd}. Agents-A1 integrates domain expertise through staged training and multi-teacher on-policy distillation \cite{agentsa12026scaling}. RL Starts before RL studies OPD as preparation for reinforcement learning, reporting higher post-RL averages than direct RL and SFT-initialized RL in text and vision-language reasoning \cite{dong2026rlstartsbeforerl}. These methods inform capability acquisition and consolidation; our training additionally conditions expert transfer on product specifications and refines the resulting policy against interaction-specific rewards.

Reward modeling determines which behavior to encourage. Process supervision evaluates intermediate reasoning \cite{lightman2023verify}, generative verifiers produce correctness judgments \cite{zhang2025genverifier}, and PoLL aggregates assessments from multiple models \cite{verga2024poll}. Aggregation can retain shared errors \cite{eisenstein2024herding}; evaluation criteria must still be defined for the target behavior. Using these assessments in optimization is a separate question. RLSD uses privileged evidence to scale token advantages while preserving their reward-derived signs \cite{yang2026selfdistilledrlvr}. DOPD routes supervision between privileged teacher and student policies to address differences in token-level supervision \cite{yu2026dopd}. DPER combines these mechanisms: it extends RLSD's magnitude modulation to dual-source evidence selected through DOPD's routing principle. Our multi-source rewards separately assess reasoning, product constraints, and goal consistency.

\section{Product-Conditioned Multimodal Data Construction}
\label{sec:data-construction}

\begin{figure}[t!]
  \centering
  \previewincludegraphics[width=\linewidth]{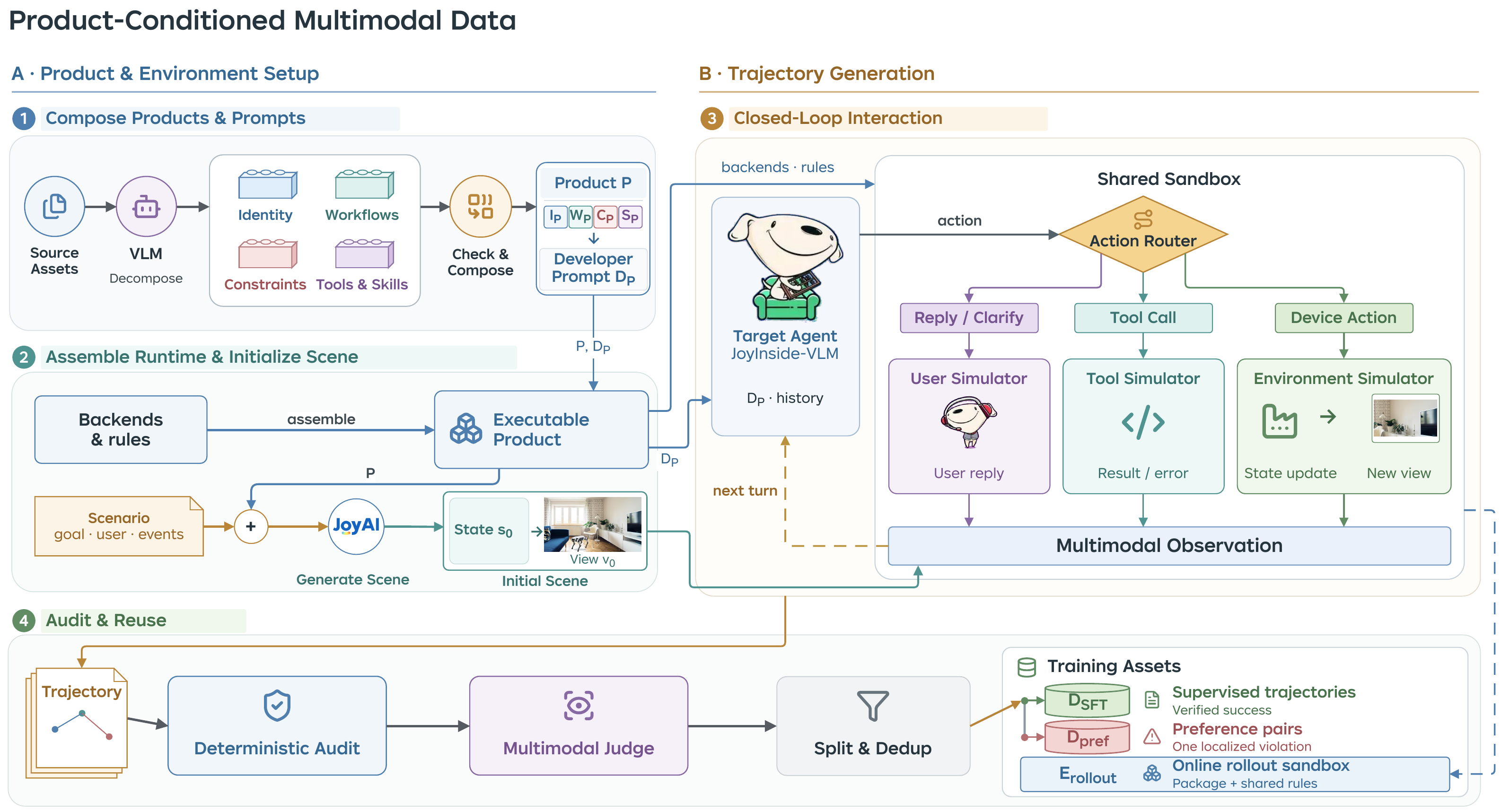}
  \caption{\textbf{Product-conditioned multimodal data construction.} \textit{(A, steps 1--2)} Compatible product modules supply the developer prompt, executable rules, and initial scene. \textit{(B, step 3)} Routed actions produce user feedback or execution results; state changes determine subsequent visual observations. \textit{(Step 4)} Deterministic and multimodal audits yield verified demonstrations and preference pairs for reward-model training. Executable environments retain the product rules for online rollouts. Verified demonstrations include appropriate refusals and compliant recovery from environmental failures as well as successful task completion.}
  \label{fig:data-pipeline}
\end{figure}

\noindent\hspace*{\parindent}We construct multimodal interaction trajectories to learn a shared policy across heterogeneous robotic platforms---including bipedal, quadruped, and wheeled robots---and smart hardware. Our construction couples compositional product specifications with executable constraints and visual feedback consistent with the resulting state. This produces interactions that cover both recovery from execution failures and adaptation to revised user goals under the same product rules. Figure~\ref{fig:data-pipeline} connects product composition and environment initialization (steps 1--2), closed-loop interaction (step 3), and trajectory validation and reuse (step 4).

\subsection{Product and Environment Construction}

\noindent\hspace*{\parindent}We use developer configurations, product descriptions, device images or interface screenshots, tool APIs, operational rules, and domain skills as source materials. JoyAI-VL performs product decomposition, code generation, trajectory generation, user simulation, and semantic auditing throughout this pipeline. It decomposes existing products into reusable modules and generates variants that preserve field restrictions and component semantics, following self-instructed data generation \cite{wang2022selfinstruct}. Each product specification is a tuple $P=(I_P,W_P,C_P,S_P)$: $I_P$ defines identity, embodiment, and interaction style; $W_P$ defines task strategies and workflows, including decision branches and recovery rules; $C_P$ specifies permissions, confirmation requirements, and operating constraints; and $S_P$ specifies tools, domain skills, and perceptual and action capabilities. Each component has a corresponding typed module library.

A code-generation agent implements simulated tool and device backends from the resource interfaces. These programs validate arguments and preconditions, execute state transitions, and return results or errors, with branches for success, invalid arguments, insufficient permission, timeout, and partial execution. Calls and state changes are logged. Before admission to a library, modules undergo interface, state-access, and failure-branch tests, together with checks that backend outputs agree with the resulting states and observations.

The composer selects modules from the four libraries and assembles compatible product configurations. Compatibility checks resolve dependencies and conflicts, ensure that workflows use installed resources and obey constraints, and require action arguments to be obtainable from user input, observations, prior results, or clarification. The selected capabilities must also match the embodiment and support the intended tasks. Joint composition allows, for example, the same workflow to operate with different embodiments, tool sets, and confirmation policies when compatible; coverage is bounded by the module libraries and compatibility rules.

Compositions that pass integration tests are compiled into a developer prompt $D_P$, resource interfaces, executable backends, state-transition rules, and observation and validation rules. Descriptive skills enter the prompt, while callable tools and device capabilities additionally expose invocation schemas and backends. The same product constraints inform both the prompt and machine-checkable predicates, aligning the behavior requested of the policy with the rules enforced and audited by the environment. Alternative prompt serializations preserve the same product semantics. Appendix~\ref{app:developer-prompt} presents a developer prompt template and its correspondence to the four product components.

For each product, we construct a user goal, user profile, initial hidden state $s_0$, visual assets, and a dynamic-event policy (Figure~\ref{fig:data-pipeline}, step 2). We specify object identities, locations, attributes, and device states before generating the initial request and image. Initial visual assets may come from real observations, simulation, or image generation, and we check them against $s_0$ before interaction begins.

\subsection{Closed-Loop Trajectory Generation}

\noindent\hspace*{\parindent}We load the product and scenario into a sandbox that maintains interaction state, a design also used in ToolSandbox \cite{lu2024toolsandbox}. In turn $t$, the agent receives the developer prompt $D_P$, prior interaction history $h_t$, and current observation $o_t=(x_t,v_t,r_t,e_t)$, where the components denote the user message, visual observation, execution result, and error, respectively. Conditioned on these inputs and the previous task state $Q_{t-1}$, it updates its task state $Q_t$ and selects an action $a_t$. Visual observations may contain images or video. Actions include responses, clarifications, refusals, tool calls, device actions, and termination. For trajectories without an explicit task queue, the policy generates only $a_t$. Hidden states, reference answers, and future event schedules are withheld from the agent.

The action router in Figure~\ref{fig:data-pipeline} (step 3) sends dialogue actions to a user simulator, as in conversational tool-use environments \cite{yao2024taubench}, while tool calls and device actions execute the registered backends. For any environment-changing call or device action, the backend returns an execution result, any error, and state changes, accounting for triggered events that affect execution. The shared sandbox applies these changes together with any remaining environmental effects, then generates visual feedback from the updated state using simulation, interface rendering, or image generation. Object, attribute, text, and multimodal semantic checks assess agreement between this feedback and the recorded state, including after failure or partial completion. User messages, execution results, errors, and images retain their separate sources in the interaction record. Appendix~\ref{app:sandbox-transition} formalizes the policy interface and this execution-to-observation sequence.

Dynamic events trigger based on specified turn, state, or action conditions. The user simulator introduces goal revisions, additional requirements, or rejected confirmations; backends report timeouts, permission changes, or partial execution; and the state engine applies changes such as object movement or device shutdown. The agent observes the consequences and must update its plan, seek clarification, retry, or terminate. These interactions provide trajectories for both recovery toward an existing goal and adaptation to a revised goal.

\subsection{Trajectory Validation and Data Outputs}

\noindent\hspace*{\parindent}An interaction ends upon goal completion, cancellation, valid refusal, unrecoverable failure, or exhaustion of the turn budget. Termination reason and task success are recorded separately. Each trajectory $\tau$ stores model-visible messages, observations, and actions alongside audit metadata: the product specification, hidden-state history, event log, and backend execution records.

Following execution-based verification in the construction of synthetic tool-use data \cite{liu2024apigen}, deterministic checks validate schemas, preconditions, permissions, confirmations, state transitions, provenance of results, and termination. A separate multimodal evaluator assesses interaction naturalness, visual grounding, adaptation to new feedback, and unsupported execution claims. Auditing uses the effective goal at each turn, including recorded user revisions. A trajectory is accepted for supervised learning only if it passes all hard checks and its semantic score reaches the acceptance threshold $\gamma$. Semantic judgments cannot override a failed hard check. Recovery demonstrations are retained only when valid actions encounter environmental failures, such as timeouts or device unavailability, and subsequent behavior remains compliant. Trajectories containing agent-side rule violations are excluded from SFT even if the agent later corrects them.

The supervised set $\mathcal{D}_{\mathrm{SFT}}$ contains accepted demonstrations, including appropriate refusals and compliant recovery behavior. We construct preference pairs by matching an accepted demonstration with an audited trajectory containing a localized violation under the same product, task, and initial conditions. Violations include missing confirmations, invalid arguments, incorrect action order, or reliance on outdated visual evidence. These pairs form $\mathcal{D}_{\mathrm{pref}}$ and supply contrasting trajectories for reward-model training (Section~\ref{sec:interaction-rewards}). The executable product, scenario, and validators form $\mathcal{E}_{\mathrm{rollout}}$, where online candidates are evaluated under the same product rules. Figure~\ref{fig:data-pipeline} (step 4) summarizes these outputs. Variants of one trajectory share a split, and deduplication uses product composition, scenario source, and visual assets.

\section{Method}
\label{sec:method}

\begin{figure}[t!]
  \centering
  \previewincludegraphics[width=0.95\linewidth]{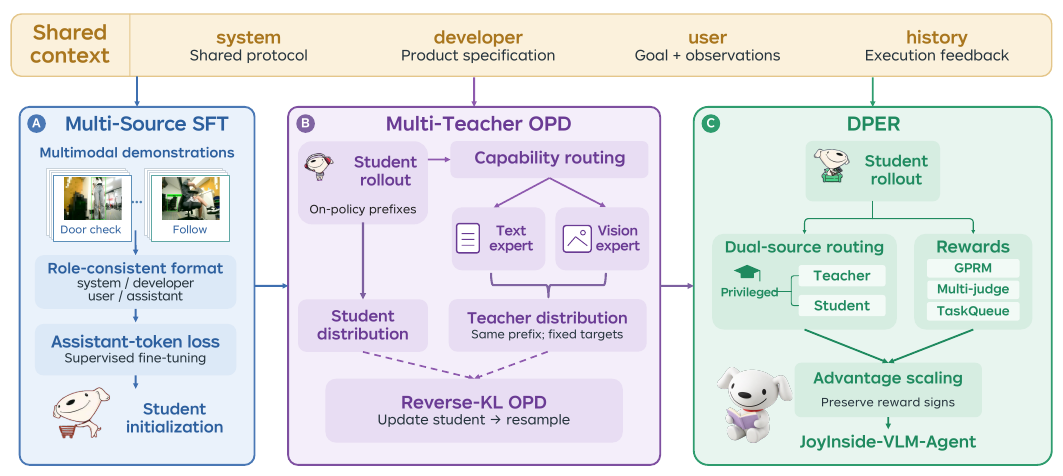}
  \caption{\textbf{Three-stage policy learning.} All stages use the shared product-conditioned context. \textit{(A)} SFT trains on assistant tokens in multimodal demonstrations. \textit{(B)} Text and vision teachers supervise student-generated prefixes through reverse-KL OPD. \textit{(C)} DPER selects privileged teacher or student evidence to rescale reward-derived token advantages while preserving their signs. Rewards combine GPRM, multi-judge rankings, and task-queue consistency. Teachers and privileged references are used only during training; deployment uses a single student.}
  \label{fig:training-framework}
\end{figure}

\noindent\hspace*{\parindent}We train a shared ReAct-style policy for bipedal, quadruped, and wheeled robots, as well as smart hardware such as interactive toys. Product specifications condition all three training stages (Figure~\ref{fig:training-framework}): supervised fine-tuning establishes interaction behavior, multi-teacher distillation transfers complementary capabilities, and DPER refines the policy using interaction rewards. Throughout training, the policy reasons and acts on the observations available at each decision. Teachers and privileged references provide training signals; deployment uses a single student with the product configuration and interaction context.

\FloatBarrier
\subsection{Product-Conditioned Supervised Fine-Tuning}
\label{sec:product-sft}

\noindent\hspace*{\parindent}Supervised fine-tuning teaches the policy to interpret product specifications and produce context-appropriate responses and tool calls. We combine general instructions, text-based tool interactions, image-conditioned decisions, multi-turn execution feedback, and verified sandbox trajectories (Section~\ref{sec:data-construction}). Following instruction tuning \cite{ouyang2022instructgpt,liu2023llava}, we organize examples from all sources using consistent message roles: system messages define the interaction protocol, developer messages encode the product specification, and user and tool messages provide requests, observations, and execution results. Our chat template serializes $D_P$ as a separate \texttt{developer} message using the same message delimiters as other roles (Appendix~\ref{app:chat-template}). Assistant messages contain the supervised reasoning, responses, and actions.

Let $\pi_\theta$ denote the student policy. For a serialized trajectory $\tau_i$, let $y_{i,n}$ be its token at position $n$ and $c_{i,n}$ the context available before that token: the protocol, product prompt $D_P$, preceding messages and assistant outputs, and available visual observations and execution feedback. Let $\mathcal{M}_i$ denote the valid assistant-token positions, excluding padding. We minimize cross-entropy over these positions, summing token losses within each trajectory and averaging over demonstrations. User messages, observations, and tool returns remain conditioning inputs, and explicit task queues, when present, are supervised as assistant outputs. Appendix~\ref{app:sft-objective} states the objective in full and specifies the mask.

The product prompt specifies the available tools, their argument requirements, and execution preconditions. Demonstrations include clarification, explanations of capability limits, and recovery from execution failures under different product configurations. Supervision therefore covers both output syntax and whether a proposed action is appropriate for the current product and observed state. The resulting checkpoint initializes multi-teacher distillation.

\paragraph{ReAct decisions and task-state generation.}
At each ReAct decision, the student generates reasoning followed by either a tool call or a direct response, such as an answer, clarification, or refusal. When required by the interaction protocol, the updated task queue is emitted in the same assistant message for either decision type. All components are generated autoregressively by the same policy and supervised as assistant outputs. The queue contains ordered batches of coarse-grained action labels, while a tool call specifies a registered tool and concrete arguments for an action in the current head batch. Tool results become observations for subsequent decisions. Neither issuing a call nor producing a direct response establishes completion of a pending tool action; completion must be verified from execution feedback. Appendix~\ref{app:joint-assistant-output} illustrates both output branches.

\subsection{Multi-Teacher On-Policy Distillation}
\label{sec:multi-teacher-opd}

\noindent\hspace*{\parindent}We use on-policy distillation \cite{agarwal2024gkd} to transfer complementary capabilities from text and vision teachers into a single student, following staged multi-teacher training \cite{agentsa12026scaling}. Text teachers supervise planning, tool selection, argument construction, and feedback interpretation; vision teachers supervise decisions requiring image evidence. Each assigned teacher receives the product constraints and student-generated prefix, together with the visual observation when required. We place this stage before DPER based on evidence that OPD can improve subsequent RL performance even when its immediate accuracy gains are small \cite{dong2026rlstartsbeforerl}.

Let $\pi_{\mathrm{old}}$ denote the student snapshot used to collect the current rollout batch, and let $\kappa(i,n)$ identify the capability-matched teacher assigned to the decision containing token $n$. We evaluate the teacher distribution $q_{i,n}(\cdot)=\pi_{T_{\kappa(i,n)}}(\cdot\mid c_{i,n})$ on the student's sampled context and stop gradients through it. With aligned teacher and student token spaces, we compute the reverse-KL OPD objective \cite{dong2026rlstartsbeforerl}, minimizing divergence from the student to the assigned teacher:
\begin{equation}
\mathcal L_{\mathrm{OPD}}(\theta)
=\mathbb E_{\tau_i\sim\pi_{\mathrm{old}}}
\left[\sum_{n\in\mathcal S_i}
D_{\mathrm{KL}}\!\left(\pi_\theta(\cdot\mid c_{i,n})\,\Vert\,q_{i,n}(\cdot)\right)\right],
\label{eq:opd-objective}
\end{equation}
where $\mathcal S_i\subseteq\mathcal M_i$ denotes the assistant positions selected for distillation. At each position, the two distributions condition on the same student-generated prefix. The expectation accounts for task sampling, student rollouts, and observations returned by the environment.

Sampled trajectories and teacher targets remain fixed during each update, and only student parameters receive gradients. We collect each subsequent rollout batch with the updated student and obtain new teacher targets. The distilled checkpoint initializes DPER; teacher assignment is used only during training.

\subsection{DPER: Token-Level Evidence Routing}
\label{sec:dper}

\noindent\hspace*{\parindent}Dual-Source Privileged On-Policy Self-Distillation with Token-Level Evidence Routing (DPER) retains RLSD's group-relative base advantage and sign-preserving advantage modulation \cite{yang2026selfdistilledrlvr}, and introduces token-level source selection inspired by DOPD \cite{yu2026dopd}. Trajectory rewards evaluate an interaction as a whole, whereas privileged references provide additional evidence about individual sampled tokens. A frozen privileged teacher and a parameter-sharing privileged student score tokens on the same rollout. Routing selects the evidence used to scale each token advantage, while the group-relative reward comparison determines its sign. DPER uses the selected evidence in a policy-gradient objective without adding DOPD's separate KL/JS distillation losses.

\subsubsection{Dual-Source Privileged Scoring}

\noindent\hspace*{\parindent}For each product-conditioned task instance, we collect a group $\mathcal C=\{\tau_i\}_{i=1}^{K}$ using $\pi_{\mathrm{old}}$. Let $z_i$ be a reference answer, solution trace, or tool trajectory available only during training. For each sampled assistant token, we compute
\begin{equation}
\begin{aligned}
p_{0,in}&=\pi_{\mathrm{old}}(y_{i,n}\mid c_{i,n}),\\
p^p_{S,in}&=\pi_{\mathrm{old}}(y_{i,n}\mid c_{i,n},z_i),\\
p^p_{T,in}&=\pi_T(y_{i,n}\mid c_{i,n},z_i),
\end{aligned}
\label{eq:dper-branch-probabilities}
\end{equation}
where $\pi_T$ is a frozen teacher with a token space aligned with the student's. The privileged student uses the same parameter snapshot as the rollout policy with the additional reference. Both privileged branches evaluate the sampled tokens under teacher forcing. Keeping the trajectory fixed makes their scores refer to the same decisions and observed consequences. The privileged student's comparison with the rollout policy changes the information available to the same model, while the teacher also contributes a different model's assessment. The context $c_{i,n}$ retains intervening user messages, observations, and tool results across ReAct turns; the generating policy never receives $z_i$.

\subsubsection{Token-Level Source Selection}

\noindent\hspace*{\parindent}We measure disagreement by $a_{in}=|\log p^p_{T,in}-\log p^p_{S,in}|$ and joint support by $u_{in}=p^p_{T,in}+p^p_{S,in}$. Their medians, $\widetilde a$ and $\widetilde u$, are computed over valid assistant tokens in the rollout batch. High and low are relative to these batch statistics. These median thresholds are our choice for adapting DOPD's source-selection principle. The selected evidence probability is
\begin{equation}
p^*_{in}
=\operatorname{sg}\!\left[
\begin{cases}
p^p_{T,in}, & a_{in}<\widetilde a\ \text{and}\ u_{in}\ge\widetilde u,\\
p^p_{T,in}, & a_{in}\ge\widetilde a\ \text{and}\ p^p_{T,in}\ge p^p_{S,in},\\
p^p_{S,in}, & \text{otherwise},
\end{cases}
\right],
\label{eq:dper-selected-probability}
\end{equation}
where $\operatorname{sg}$ stops gradients through the selected score. The conditions define four cases, following DOPD's source-selection structure \cite{yu2026dopd}:
\begin{itemize}[leftmargin=1.5em,label=\textbullet]
  \item Low disagreement, high joint support (LH). The two sources agree with relatively high joint support for the sampled token, and we select the privileged teacher.

  \item Low disagreement, low joint support (LL). Agreement is accompanied by low joint support for the sampled token, and we select the privileged student.

  \item High disagreement, stronger teacher support (HT). We select the privileged teacher when its probability for the sampled token is at least as high as the student's, including ties.

  \item High disagreement, stronger student support (HS). We select the privileged student when its probability for the sampled token exceeds the teacher's.
\end{itemize}

All four cases use the same advantage-modulation rule below. Probability support alone does not establish that an action is executable or consistent with the user's goal.

\subsubsection{Evidence-Weighted Policy Optimization}

\noindent\hspace*{\parindent}Following RLSD \cite{yang2026selfdistilledrlvr}, we compute group-relative base advantages $A_i$ from the composite trajectory rewards in Equation~\eqref{eq:reward-total} within each same-task rollout group. We retain RLSD's evidence-weighting rule, using the routed probability $p^*_{in}$ as its privileged score:
\begin{equation}
\Delta_{in}=\operatorname{sg}\!\left[\log p^*_{in}-\log p_{0,in}\right],
\qquad
w_{in}=\exp\!\left(\operatorname{sign}(A_i)\Delta_{in}\right).
\label{eq:dper-evidence-weight}
\end{equation}
The token advantage is
\begin{equation}
\widehat A_{in}
=A_i\left[(1-\lambda)+\lambda\operatorname{clip}(w_{in},1-\epsilon_w,1+\epsilon_w)\right],
\label{eq:dper-token-advantage}
\end{equation}
where $\lambda\in[0,1]$ controls evidence strength and $\epsilon_w\ge0$ is the clipping radius. The multiplier is positive, preserving the sign of each nonzero advantage and leaving zero advantages unchanged. Relative to the rollout policy, greater privileged support increases reinforcement for positive advantages, while weaker support increases suppression for negative advantages, subject to clipping. Clipping bounds the influence of large probability contrasts, and mixing with the original advantage controls the contribution of privileged evidence. When $\lambda=0$ or $\epsilon_w=0$, the original trajectory advantage is recovered. A selected probability equal to the rollout probability also leaves the corresponding token advantage unchanged. The sign guarantee applies to individual token advantages. If routing always selects $p^p_{S,in}$, the token-advantage formula reduces to RLSD's self-distillation modulation under the same rewards and hyperparameters.

We maximize a clipped policy surrogate using the evidence-weighted token advantages, averaging over valid assistant tokens within each trajectory and then over task instances and their sampled trajectories. Appendix~\ref{app:dper-objective} gives the full objective and policy-ratio definition. Privileged scores, routing decisions, and token advantages are computed once per rollout batch, detached from gradients, and held fixed during its policy updates. They are recomputed from the next student snapshot after new rollouts are collected. Only the unprivileged policy is optimized for deployment.

\subsection{Multi-Source Interaction Rewards}
\label{sec:interaction-rewards}

\noindent\hspace*{\parindent}We evaluate process quality through GPRM, product-constrained behavior through multi-judge ranking, and goal-consistent planning through TaskQueue. For the queue-bearing interaction path described here, normal reward aggregation requires a successfully constructed scoring input and a valid GPRM score, including zero. Let $\mathcal B_i\subseteq\{\mathrm{GPRM},\mathrm{Judge},\mathrm{Queue}\}$ contain GPRM and the available optional scores for candidate $\tau_i$. Its normal-path reward is
\begin{equation}
R_{\mathrm{content},i}
=\frac{\sum_{b\in\mathcal B_i}\beta_b R_{b,i}}
{\sum_{b\in\mathcal B_i}\beta_b},
\qquad
R_i=R_{\mathrm{content},i}\,\ell_{\mathrm{format},i}\,\ell_{\mathrm{len},i},
\label{eq:reward-total}
\end{equation}
where $\beta_b\ge0$ and $\beta_{\mathrm{GPRM}}>0$, ensuring a positive denominator. The multiplicative factors $\ell_{\mathrm{format},i}\in[0.05,1]$ and $\ell_{\mathrm{len},i}\in[0.1,1]$ assess output structure and length. Missing optional scores are excluded and the remaining weights are renormalized; valid zero scores from any branch retain their weights.

The format factor assesses compliance with the required output structure. The interaction protocol determines whether a task queue is required; choosing a direct response does not waive an existing queue requirement. Dialogue-only outputs for which the protocol requires no queue incur no missing-queue penalty. Empty outputs receive the minimum coefficient. An otherwise usable output that omits a required task queue receives a coefficient above this minimum but below that of a complete valid output. If this omission makes the Queue score unavailable on the normal scoring path, the coefficient is capped at the ratio of the remaining available reward weight to that same weight plus the Queue weight. With other scores and the length factor held fixed, this cap prevents renormalization alone from increasing the reward relative to a valid output with a zero Queue score. The length factor penalizes excessive reasoning and response content using task-dependent allowances.

If the scoring input cannot be constructed, the scorer returns $R_i=0$. If GPRM is unavailable because no process step can be evaluated, its request fails, or no valid score is returned, the scorer instead assigns $R_i=0.1\,\ell_{\mathrm{format},i}\ell_{\mathrm{len},i}$ and skips Judge and Queue evaluation for that candidate. A valid zero GPRM score follows normal aggregation and does not trigger this fallback. This fixed fallback is an implementation rule, not a process-quality judgment. These candidate-level rules precede normal aggregation; group-level exclusions are described below. Appendix~\ref{app:reward-implementation} details score availability and fallback handling. The resulting rewards supply DPER's group-relative advantages.

\subsubsection{Generative Process Evaluation}

\noindent\hspace*{\parindent}GPRM combines process supervision \cite{lightman2023verify} with generative verification \cite{zhang2025genverifier}. For each reasoning step $s$, its input contains the task and trajectory prefix through that step. Prefix evaluation exposes intermediate errors that a final-answer judgment does not separately represent. The evaluator generates a verification rationale, then predicts an ordered quality grade $g\in\mathcal G=\{A,B,C,D\}$ at a fixed judgment position. The rationale assesses correctness and contextual relevance and identifies errors in the current reasoning. This evaluation uses the language model's output distribution without an additional scoring head.

We obtain grade probabilities $p_{s,g}$ by applying softmax only over the four grade-token logits after the rationale (Appendix~\ref{app:gprm-normalization}). With grade weights $(w_A,w_B,w_C,w_D)=(1,\frac23,\frac13,0)$, the step reward and trajectory score are
\begin{equation}
r_s^{\mathrm{proc}}=\sum_{g\in\mathcal G}w_g p_{s,g},
\qquad
R_{\mathrm{GPRM},i}=\frac{1}{L_i}\sum_{s=1}^{L_i}r_s^{\mathrm{proc}},
\label{eq:gprm-trajectory-reward}
\end{equation}
where $L_i>0$ is the number of evaluated reasoning steps; the candidate index is suppressed on step-level quantities. With no evaluable step, the scorer takes the fallback above without evaluating this average. Grades run from highest quality (A) to lowest (D). The expectation retains uncertainty across grades, and averaging puts trajectories with different numbers of evaluated steps on the same score range, $[0,1]$. The result measures expected process quality under the evaluator's grade distribution.

The evaluator's training data include trajectories from the preference pairs in $\mathcal{D}_{\mathrm{pref}}$ (Section~\ref{sec:data-construction}). It is trained with autoregressive cross-entropy on annotated sequences containing the task, reasoning prefix, verification rationale, and quality grade. Rationale annotations record the judgment basis, identified errors, and their implications for subsequent reasoning. These labels supervise process quality independently of the candidate rankings below.

\subsubsection{Multi-Judge Ranking}

\noindent\hspace*{\parindent}Each judge receives the same complete candidate group $\mathcal C$ and produces a tiered ranking with ties. This relative protocol avoids requiring scores from different judges to share an absolute scale \cite{niu2026absolutetorelative}. Task-dependent criteria cover product capabilities, behavioral constraints, role consistency, planning dependencies, use of observations, goal completion, and redundancy within and across candidates. Presenting the full group lets judges compare alternative behaviors under the same task and identify repeated content across rollouts. Their outputs are ranking ballots over shared candidate identities.

For a complete ranking group with $K>1$ and at least one valid ballot, let $\bar r_{mi}$ be candidate $i$'s average rank in valid ballot $m$, with rank 1 best. Tied candidates share the mean of their occupied ranks. Invalid ballots are discarded in full; the remaining ballots form $\mathcal V$, with nonnegative weights renormalized so that $\sum_{m\in\mathcal V}\alpha_m=1$. We aggregate their normalized rank scores into $b_i$, rerank candidates by decreasing $b_i$, and map the resulting average rank $\bar r_i$ to the final reward:
\begin{equation}
b_i=\sum_{m\in\mathcal V}\alpha_m\frac{K-\bar r_{mi}}{K-1},
\qquad
R_{\mathrm{Judge},i}=\delta_{\mathrm J}+(1-\delta_{\mathrm J})\frac{K-\bar r_i}{K-1},
\label{eq:judge-ranking-reward}
\end{equation}
where $\delta_{\mathrm J}$ is a tunable reward-calibration hyperparameter, with its value specified in Section~\ref{sec:experimental-setup}. Both ranking stages preserve ties. Weighted ballot scores determine the aggregate ordering; the final mapping uses rank spacing rather than numerical gaps between aggregate scores. These scores express relative preferences within the group; they do not measure absolute task success or define GPRM grades. If every candidate in a group triggers a deterministic hard failure, we skip both relative ranking and DPER policy updates for that group. If no valid ballot remains, the Judge branch is unavailable: normal-path candidates use the remaining scores in Equation~\eqref{eq:reward-total}, while candidates assigned an early-exit or fallback reward retain that value.

\subsubsection{Task-Queue Consistency}

\noindent\hspace*{\parindent}TaskQueue evaluates semantic alignment and continuity across turns. At turn $r$, let $Q^{(r)}=[B_1^{(r)},\ldots,B_{L_r}^{(r)}]$, with the head batch representing the current tasks. On the first turn, JoyAI-VL inversely rewrites the queue as a natural-language task description under task-granularity rules. The first-turn score compares this reconstructed intent with the initial request $x$, as defined in Equation~\eqref{eq:taskqueue-first-turn-reward}. It evaluates intent coverage without requiring an exact match to a reference queue and uses no continuity term.

Queues use nested batches of coarse-grained action labels without concrete objects or arguments. The grouping rules separate actions from queries, place individual queries or broadcasts in separate batches, and start a new batch when the interaction subject changes. These rules specify the queue representation; inverse rewriting preserves the proposed tasks and grouping without adding, removing, or correcting them.

On feedback turns with a nonempty observation block and a preceding queue, $q_{\mathrm{obs}}^{(r)}$ compares a JoyAI-VL inverse rewrite of the current queue with task requirements updated from the observation history, including user revisions, visual observations, execution feedback, and errors. Qwen3-Embedding-4B encodes the compared texts, and cosine similarity is mapped to $[0,1]$, consistent with the overall reward range. The continuity term $q_{\mathrm{cont}}^{(r)}$ compares the queue against the preceding queue updated by verified execution: completed heads are removed, failed or unexecuted heads are retained, and partially completed heads retain their unfinished tasks in order. It counts exact batch matches at corresponding positions and normalizes by the longer queue's length. A batch matches only if all task entries are equal at the same positions; neither batches nor tasks within a batch may be reordered. With both scores available,
\begin{equation}
q_{\mathrm{queue}}^{(r)}
=\frac{\eta_{\mathrm{obs}}q_{\mathrm{obs}}^{(r)}
+\eta_{\mathrm{cont}}q_{\mathrm{cont}}^{(r)}}
{\eta_{\mathrm{obs}}+\eta_{\mathrm{cont}}},
\label{eq:taskqueue-combined-reward}
\end{equation}
with nonnegative weights and a positive denominator. All components lie in $[0,1]$. Available components retain their weights, which are renormalized when another component is missing; evaluated zeros remain included. If no component is available, the turn has no Queue score. The two terms balance adaptation to revised goals and preservation of pending task order: a justified revision may reduce positional agreement, so exact queue matching alone is insufficient. The trajectory score $R_{\mathrm{Queue},i}$ averages evaluated turn scores, including the initial score; it is unavailable if no turn is evaluable. Appendix~\ref{app:taskqueue-semantics} gives the component definitions and empty-queue handling, and Table~\ref{tab:taskqueue-inverse-prompt} provides the inverse-rewriting prompt.

\section{Experiments}
\label{sec:experiments}

\noindent\hspace*{\parindent}We evaluate JoyInside-VLM-Agent on four public tool-use benchmarks and two proprietary evaluation sets, analyze training dynamics, and examine two physical-robot interactions. The quantitative results assess tool-call trajectories, step-wise decisions, argument matching, and response overlap; the cases illustrate decisions based on visual evidence and execution feedback.

\subsection{Experimental Setup}
\label{sec:experimental-setup}

\paragraph{Training data.}
Our training corpus comprises approximately 120K examples: 50K proprietary interaction trajectories, 40K supplementary examples adapted from public agent datasets, and 30K video-instruction examples from LLaVA-Video-178K \cite{zhang2024llavavideo} and Video-ChatGPT \cite{maaz2023videochatgpt}. The proprietary trajectories span quadruped, bipedal, and wheeled robots and AI-enabled plush toys. The supplementary subset draws on four public sources: $\tau^2$-Bench \cite{barres2025tau2bench} for conversational tool use and agent--user coordination, VitaBench \cite{he2025vitabench} for interactive tasks spanning multiple service scenarios, DeepPlanning \cite{zhang2026deepplanning} for long-horizon planning with verifiable constraints, and SWE-bench Verified \cite{openai2024swebenchverified} for repository-level issue resolution. We refine and augment all four sources using the task-variation, interaction-generation, and validation procedures adapted from Section~\ref{sec:data-construction}, producing supplementary training data for our model. The 40K count refers to the examples retained after augmentation and filtering.

\paragraph{Evaluation datasets.}
We evaluate on MSAgent \cite{msagent2023}, BFCL v4 \cite{yan2024bfcl}, ToolBench \cite{qin2023toolllm}, and ToolACE \cite{liu2024toolace}. We report benchmark-level scores on all four datasets and additional step-wise and response-overlap metrics on ToolBench and ToolACE. We also evaluate on two proprietary JoyInside datasets. JoyInside-Devices focuses on extended multi-turn dialogue and simple device control across AI-enabled toys, smart appliances, and other AIoT devices. Tasks span conversational assistance, context tracking, evolving user preferences, status queries, and basic control commands. JoyInside-Robotics focuses on long-horizon, visually grounded reasoning and complex robot control. Tasks involve target search and localization, spatial reasoning, multi-step planning and execution, and adaptive replanning based on visual observations and execution feedback.

\paragraph{Models and product configuration.}
JoyInside-VLM-Agent starts from Qwen3.5-35B-A3B and undergoes supervised fine-tuning, multi-teacher distillation, and DPER (Section~\ref{sec:method}). We compare the resulting student with the base model and Qwen3.5-397B-A17B. At inference, it uses product specifications and interaction history; privileged references are limited to training. We also use SkillOpt \cite{yang2026skillopt} to refine the domain-skill instructions used on JoyInside-Devices and JoyInside-Robotics while keeping model parameters fixed during skill refinement. Results on these sets therefore reflect the trained policy and refined skills together.

\paragraph{Training teachers.}
For the experiments reported here, we use Qwen3.8-27B for the OPD teachers and the frozen DPER teacher. Teacher selection is configurable, subject to the token-space alignment required by the objectives in Section~\ref{sec:method}.

\paragraph{DPER training configuration.}
We train DPER on 16 NVIDIA H200 GPUs for 350 steps with 128 prompts per batch and $K=8$ responses per prompt, yielding 1,024 sampled responses. The prompt minibatch size is 32, with an actor microbatch size of one sequence per GPU. The learning rate is $5\times10^{-6}$, with 10 warmup steps and weight decay 0.1. Rollouts use temperature 1.0, top-$p$ 0.95, and top-$k$ 20, with maximum prompt and response lengths of 16,384 and 2,048 tokens, respectively. DPER uses evidence strength $\lambda=0.5$. Policy losses are averaged over valid tokens within each response and then over responses. Privileged teacher and student branches provide scores only; separate distillation loss, KL regularization, and the entropy bonus are disabled. In Equation~\eqref{eq:reward-total}, the GPRM, Judge, and Queue weights are 0.4, 0.3, and 0.3, respectively, renormalized over available scores; the fallback content factor is 0.1. We set the Judge calibration hyperparameter in Equation~\eqref{eq:judge-ranking-reward} to $\delta_{\mathrm J}=0.3$. Within TaskQueue, Equation~\eqref{eq:taskqueue-combined-reward} uses $\eta_{\mathrm{obs}}=0.6$ for semantic agreement and $\eta_{\mathrm{cont}}=0.4$ for continuity. Format and length factors penalize structural violations and excessive output length, with length allowances adjusted to task complexity (Section~\ref{sec:interaction-rewards}).

\paragraph{Benchmark-level evaluation.}
For MSAgent, ToolBench, and ToolACE, we report trajectory accuracy, measured by exact match of the complete tool-call sequence. For BFCL v4, we use the official evaluation script and report its overall score, following the benchmark's category-specific evaluation and aggregation rules. Table~\ref{tab:benchmark-results}(a) reports the resulting scores on a 0--100 scale.

\paragraph{Step-wise and response metrics.}
For ToolBench, ToolACE, JoyInside-Devices, and JoyInside-Robotics, Plan.EM measures decision-type exact match at evaluated steps, rather than the correctness of a complete plan. Act.EM measures tool-name exact match, and Avg.F1 gives partial credit for argument matching. HalluRate is the percentage of evaluated tool-call records that select a tool outside the available capability set; it measures unsupported tool selection. ROUGE-L \cite{lin2004rouge} measures lexical overlap between final and reference responses. All scores in Table~\ref{tab:benchmark-results} are reported on a 0--100 scale; only HalluRate is better when lower.

\subsection{Training Dynamics}
\label{sec:training-dynamics}
\label{sec:algorithm-comparisons}

\begin{figure}[!htbp]
  \centering
  \includegraphics[width=\linewidth]{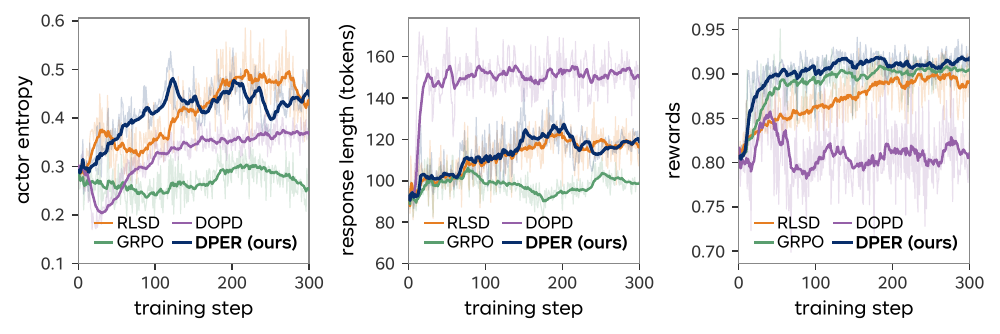}\par
  \vspace{2pt}
  {\jdaffiliationfamily\small\noindent
  \makebox[\dimexpr\linewidth/3\relax][c]{(a) Entropy}%
  \makebox[\dimexpr\linewidth/3\relax][c]{(b) Response length}%
  \makebox[\dimexpr\linewidth/3\relax][c]{(c) Maximum reward}%
  \par}
  \caption{\textbf{Training dynamics over the first 300 steps.} DPER is compared with GRPO, RLSD, and DOPD in terms of (a) actor entropy, (b) response length in tokens, and (c) maximum reward. Dark curves show smoothed trends; faint traces show step-to-step fluctuations.}
  \label{fig:training-dynamics}
\end{figure}

\noindent\hspace*{\parindent}Figure~\ref{fig:training-dynamics} compares recorded training runs of DPER, GRPO, RLSD \cite{yang2026selfdistilledrlvr}, and DOPD \cite{yu2026dopd} over the first 300 steps. We track actor entropy, response length, and maximum reward throughout optimization. The DPER run uses the configuration in Section~\ref{sec:experimental-setup} and continues to 350 steps.

DPER's maximum reward rises rapidly and remains above the baselines for most plotted steps (panel c), while response length grows moderately before fluctuating near RLSD and below DOPD (panel b). Its entropy rises early and then fluctuates around 0.40--0.48, generally above GRPO but sometimes below RLSD (panel a). Maximum reward characterizes the highest-scoring sampled output and does not establish performance across all rollouts. These curves describe training behavior; held-out evaluations and component ablations are needed to assess generalization and isolate routing and reward contributions.

\FloatBarrier
\subsection{Benchmark Results}
\label{sec:benchmark-results}

\noindent\hspace*{\parindent}Table~\ref{tab:benchmark-results} reports benchmark-level scores on four public datasets and step-wise and response-overlap metrics on ToolBench, ToolACE, and the two proprietary sets.

\begin{table}[t!]
  \centering
  \caption{Tool-use performance on public benchmarks and proprietary evaluation sets. Boldface indicates the best result for each dataset and metric.}
  \label{tab:benchmark-results}
  \small
  \setlength{\tabcolsep}{5pt}
  \renewcommand{\arraystretch}{1.12}
  \begin{tabular}{@{}lrrr@{}}
    \toprule
    \multicolumn{4}{@{}l}{\textbf{(a) Benchmark-level scores} ($\uparrow$)} \\
    \midrule
    Dataset & \multicolumn{1}{c}{Qwen3.5-35B-A3B}
            & \multicolumn{1}{c}{Qwen3.5-397B-A17B}
            & \multicolumn{1}{c}{JoyInside-VLM-Agent} \\
    \midrule
    MSAgent & 88.46 & \textbf{92.41} & 90.39 \\
    BFCL v4 & 67.30 & 72.90 & \textbf{77.44} \\
    ToolBench & 18.30 & 21.00 & \textbf{23.50} \\
    ToolACE & 50.92 & \textbf{52.27} & 51.53 \\
    \bottomrule
  \end{tabular}

  \medskip
  \small
  \renewcommand{\arraystretch}{1.20}
  \setlength{\tabcolsep}{3pt}
  \begin{tabular}{@{}llrrrrr@{}}
    \toprule
    \multicolumn{7}{@{}l}{\textbf{(b) Step-wise tool-use and response metrics}} \\
    \midrule
    Dataset & Model & Plan.EM$\uparrow$ & Act.EM$\uparrow$
            & Avg.F1$\uparrow$ & HalluRate$\downarrow$
            & ROUGE-L$\uparrow$ \\
    \midrule
    \multicolumn{7}{@{}l}{\textit{Public benchmarks}} \\
    ToolBench & Qwen3.5-35B-A3B & 74.48 & 32.72 & 27.94 & 1.14 & 3.92 \\
     & Qwen3.5-397B-A17B & 80.77 & 43.55 & 38.78 & \textbf{0.31} & 5.11 \\
     & JoyInside-VLM-Agent & \textbf{81.03} & \textbf{54.66} & \textbf{49.17} & 0.52 & \textbf{11.84} \\
    \addlinespace[3pt]
    ToolACE & Qwen3.5-35B-A3B & 96.16 & 88.19 & 75.59 & 0.70 & 35.45 \\
     & Qwen3.5-397B-A17B & 96.70 & 88.63 & \textbf{75.91} & \textbf{0.44} & \textbf{37.72} \\
     & JoyInside-VLM-Agent & \textbf{96.86} & \textbf{89.06} & 75.71 & 0.52 & 35.24 \\
    \midrule
    \multicolumn{7}{@{}l}{\textit{Proprietary evaluation sets}} \\
    JoyInside-Devices & Qwen3.5-35B-A3B & 71.63 & 45.44 & 33.47 & 0.19 & 13.02 \\
     & Qwen3.5-397B-A17B & 75.64 & 60.04 & 45.17 & 0.10 & 25.48 \\
     & JoyInside-VLM-Agent & \textbf{89.90} & \textbf{84.99} & \textbf{57.18} & \textbf{0.02} & \textbf{26.57} \\
    \addlinespace[3pt]
    JoyInside-Robotics & Qwen3.5-35B-A3B & 92.76 & 87.85 & 74.84 & 0.43 & 9.76 \\
     & Qwen3.5-397B-A17B & 94.10 & 90.56 & 77.89 & 0.25 & 12.04 \\
     & JoyInside-VLM-Agent & \textbf{97.58} & \textbf{93.73} & \textbf{83.10} & \textbf{0.04} & \textbf{23.30} \\
    \bottomrule
  \end{tabular}

  \medskip
  \parbox{.72\linewidth}{\raggedright\footnotesize\textit{Note.} Scores are reported on a 0--100 scale; arrows indicate the preferred direction. Panel (a) reports trajectory accuracy for MSAgent, ToolBench, and ToolACE, and the BFCL v4 overall score. Boldface marks the best result within each dataset and metric. JoyInside-VLM-Agent is initialized from Qwen3.5-35B-A3B. Trajectory accuracy was not reported for JoyInside-Devices or JoyInside-Robotics. Results for our model on the proprietary evaluation sets include domain-specific skill refinement and therefore reflect the evaluated system configuration.}
\end{table}
\paragraph{Public benchmarks.}
JoyInside-VLM-Agent achieves a BFCL v4 overall score of 77.44, exceeding Qwen3.5-397B-A17B and Qwen3.5-35B-A3B by 4.54 and 10.14 points, respectively. On MSAgent, it reaches 90.39\% trajectory accuracy, between the larger (92.41\%) and base (88.46\%) models.

On ToolBench, JoyInside-VLM-Agent leads in trajectory accuracy (23.5\%) and four of the five tool-use and response metrics. Its Act.EM of 54.66 exceeds that of the larger model by 11.11 points, although the latter has a lower HalluRate (0.31 versus 0.52). On ToolACE, our model has the highest Plan.EM (96.86) and Act.EM (89.06), but trails the larger model in trajectory accuracy (51.53\% versus 52.27\%), Avg.F1, HalluRate, and ROUGE-L. Thus, gains in decision-type and tool-name matching do not consistently extend to complete-trajectory accuracy or argument matching.

\paragraph{Proprietary datasets.}
JoyInside-VLM-Agent leads on all five reported metrics for both JoyInside-Devices and JoyInside-Robotics. Its Act.EM reaches 84.99 and 93.73, exceeding Qwen3.5-397B-A17B by 24.95 and 3.17 points, respectively; Avg.F1 reaches 57.18 and 83.10, while HalluRate falls to 0.02\% and 0.04\%. These results characterize the trained policy with refined domain skills. We do not report trajectory exact-match accuracy for these two sets.

\subsection{Real-World Case Studies}
\label{sec:real-world-cases}

\noindent\hspace*{\parindent}Two office-based ReAct trajectories on a quadruped illustrate how visual evidence and execution feedback guide a greeting task: in one, the target is initially outside the camera view; in the other, the target disappears after an attempted approach.

\begin{figure}[htbp]
  \centering
  \previewincludegraphics[width=\linewidth]{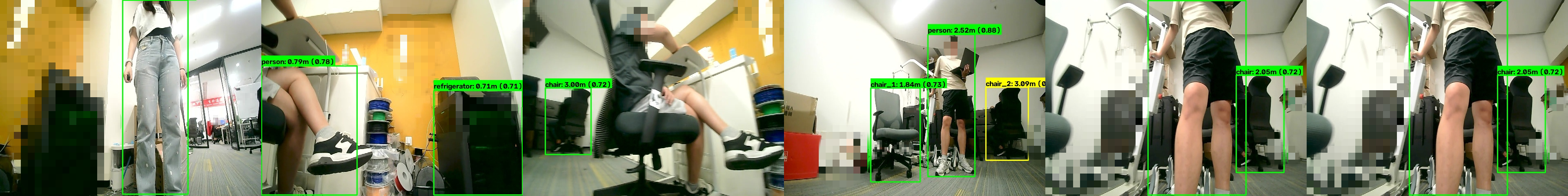}\par
  \vspace{2pt}
  {\jdaffiliationfamily\small\noindent
  \makebox[\dimexpr\linewidth/6\relax][c]{Step~1}%
  \makebox[\dimexpr\linewidth/6\relax][c]{Step~2}%
  \makebox[\dimexpr\linewidth/6\relax][c]{Step~3}%
  \makebox[\dimexpr\linewidth/6\relax][c]{Step~4}%
  \makebox[\dimexpr\linewidth/6\relax][c]{Step~5}%
  \makebox[\dimexpr\linewidth/6\relax][c]{Step~6}%
  \par}
  \caption{\textbf{Visual search before greeting.} The requested person is outside the quadruped's camera view in steps 1--3, identified by clothing and laptop attributes in step 4, and seen at close range in steps 5--6. Frames follow the interaction from left to right; bounding boxes show object detections.}
  \label{fig:case-search}
\end{figure}

\paragraph{Greeting under partial observability.}
The robot is asked to greet a man in a light shirt and black shorts who holds a laptop (Figure~\ref{fig:case-search}). Because the target is not initially visible in the camera view, it searches with \texttt{turn\_left} and checks each observation against the requested attributes. After detecting a match, it calls \texttt{visual\_approach}, then \texttt{perform\_greeting}, and reports completion. The transition from search to approach follows new visual evidence.

\begin{figure}[htbp]
  \centering
  \previewincludegraphics[width=\linewidth]{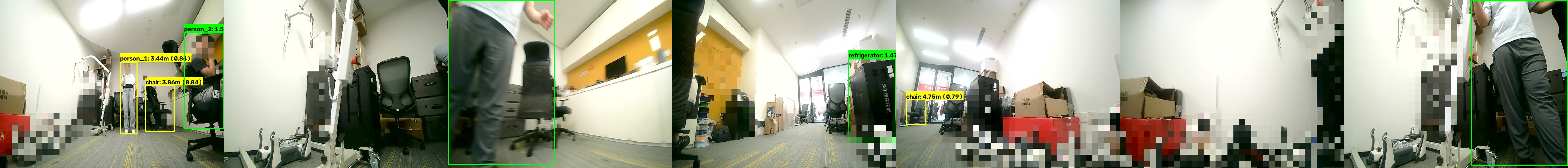}\par
  \vspace{2pt}
  {\jdaffiliationfamily\small\noindent
  \makebox[\dimexpr\linewidth/7\relax][c]{Step~1}%
  \makebox[\dimexpr\linewidth/7\relax][c]{Step~2}%
  \makebox[\dimexpr\linewidth/7\relax][c]{Step~3}%
  \makebox[\dimexpr\linewidth/7\relax][c]{Step~4}%
  \makebox[\dimexpr\linewidth/7\relax][c]{Step~5}%
  \makebox[\dimexpr\linewidth/7\relax][c]{Step~6}%
  \makebox[\dimexpr\linewidth/7\relax][c]{Step~7}%
  \par}
  \caption{\textbf{Visual search after a failed approach.} The target is visible in step 1 but is not reacquired during the quadruped's subsequent rotational search in steps 2--7. Frames follow the interaction from left to right; bounding boxes show object detections.}
  \label{fig:case-recovery}
\end{figure}

\FloatBarrier
\paragraph{Replanning after a failed approach.}
The robot is asked to greet a woman in a white shirt and light blue jeans (Figure~\ref{fig:case-recovery}). Its initial \texttt{visual\_approach} fails and the target leaves the field of view. The robot then searches with \texttt{turn\_right}, rejects mismatched people, and reports failure after six subsequent search steps without reacquiring the target. Execution feedback thus triggers replanning, and the exhausted search budget triggers termination.

These cases illustrate visually grounded decisions, feedback-driven replanning, and explicit reporting of success or failure. The quadruped trajectories are selected qualitative examples within a broader application scope that also includes humanoid robots and AI devices.

\section{Conclusion}
\begingroup
\setlength{\emergencystretch}{3em}
We presented JoyInside-VLM-Agent, a multimodal agent that combines verified, product-conditioned trajectories with staged policy learning to adapt behavior to the roles, capabilities, and constraints of robotic platforms and smart hardware. Multi-source rewards assess reasoning quality, adherence to product constraints, and alignment with evolving goals. DPER routes token-level evidence between privileged teacher and student policies to scale token advantages while preserving their reward-derived signs. Built on Qwen3.5-35B-A3B, the model achieves a BFCL v4 overall score of 77.44, while results on the other public benchmarks are mixed. On JoyInside-Devices and JoyInside-Robotics, the evaluated system, which includes domain-specific skill refinement, leads the baselines on all reported metrics. Two quadruped case studies illustrate visually grounded search, replanning from execution feedback, and explicit failure reporting.
\par
\endgroup

\section{Limitations and Future Work}
\label{sec:limitations}

\noindent\hspace*{\parindent}JoyInside-VLM-Agent has been applied to quadruped robots, humanoid robots, and AI devices. Generalization to a broader range of product configurations and operating environments still requires systematic evaluation. Product-conditioned data construction also depends on the coverage of the capability library and the fidelity of sandbox-generated feedback. Future work will expand product coverage and measure task completion and recovery over longer real-world interactions. We also plan to evaluate across more diverse benchmarks in multimodal understanding, reasoning, planning, and tool use to assess the generality of the learned capabilities beyond the current evaluation suite.

The current results do not isolate the contributions of data construction, reward design, and DPER routing, and the proprietary evaluations additionally include domain-specific skill refinement. Controlled ablations are needed to separate these effects. Moreover, model-based rewards and embedding similarity provide imperfect proxies for task success; preserving reward-derived advantage signs does not resolve errors in the underlying reward. We plan to investigate execution-grounded reward calibration and robustness to incomplete or noisy feedback, alongside the component ablations, to better understand how each stage supports reliable product-specific behavior.

\bibliographystyle{unsrtnat}
\bibliography{references}

\clearpage
\appendix
\numberwithin{equation}{section}

\section{Supplementary Formulations}
\label{app:formulations}

\subsection{Policy Decisions and Sandbox Transitions}
\label{app:sandbox-transition}

\noindent\hspace*{\parindent}Let $\pi_{\mathrm{gen}}$ denote the trajectory-generation policy used for offline data construction in Section~\ref{sec:data-construction}. At turn $t$, it conditions on the product prompt $D_P$, prior interaction history $h_t$, current observation $o_t=(x_t,v_t,r_t,e_t)$, and previous task state $Q_{t-1}$. It generates an updated task state and an action:
\begin{equation}
(Q_t,a_t)\sim \pi_{\mathrm{gen}}(\cdot\mid D_P,h_t,o_t,Q_{t-1}),
\label{eq:data-policy}
\end{equation}
where $h_t$ excludes $o_t$; the history and previous queue are empty before the first decision. Observation components follow Section~\ref{sec:data-construction}, with $\varnothing$ denoting absence. Queue variables are omitted when unused. Online rollouts substitute the current student for $\pi_{\mathrm{gen}}$.

For product $P$, the backend $\mathcal B_P$ executes action $a_t$ in hidden state $s_t$, accounting for triggered events $\epsilon_t$ and their timing. It returns the result $r_{t+1}$, error $e_{t+1}$, and state changes $\Delta s_t$; $F_P$ applies these changes and any remaining event effects. When visual feedback is produced,
\begin{equation}
\begin{aligned}
(r_{t+1},e_{t+1},\Delta s_t) &= \mathcal{B}_P(a_t,s_t,\epsilon_t),\\
s_{t+1} &= F_P(s_t,a_t,\Delta s_t,\epsilon_t),\\
v_{t+1} &= G_{\mathrm{vis}}(v_t,s_t,s_{t+1}).
\end{aligned}
\label{eq:sandbox-transition}
\end{equation}
The visual generator uses simulation, interface rendering, or image generation; its output is checked against $s_{t+1}$. Dialogue replies $x_{t+1}$ come from the user simulator, while read-only calls cause no action-induced state changes. These branches need not produce visual feedback. Hidden states and event schedules remain unavailable to the policy.

\subsection{Supervised Learning Objective}
\label{app:sft-objective}

\noindent\hspace*{\parindent}Using the notation in Section~\ref{sec:product-sft}, we minimize the average negative log-likelihood of the supervised assistant tokens:
\begin{equation}
\label{eq:sft-objective}
\mathcal{L}_{\mathrm{SFT}}(\theta) = \mathbb{E}_{\tau_i \sim D_{\mathrm{SFT}}} \left[ \sum_{n \in \mathcal{M}_i} -\log \pi_\theta \left( y_{i,n} \mid c_{i,n} \right) \right].
\end{equation}
The mask $\mathcal{M}_i$ includes assistant reasoning, responses, actions, and explicit task queues when present, while excluding padding. Structural delimiters such as tags and brackets are supervised together with the surrounding content tokens and carry no separate weight. User messages, observations, and tool returns contribute to $c_{i,n}$ but are not prediction targets. Equation~\eqref{eq:sft-objective} sums over the supervised positions of each trajectory and then averages over $D_{\mathrm{SFT}}$, so a demonstration with more supervised tokens contributes proportionally more. We keep this convention for the token-level objectives in Section~\ref{sec:product-sft} and Section~\ref{sec:multi-teacher-opd}, and average within each trajectory for the DPER policy loss (Appendix~\ref{app:dper-objective}), which weights a single trajectory-level advantage: summing over tokens there would multiply each update by the length of its trajectory and let long trajectories dominate the policy gradient.

\subsection{DPER Policy Objective}
\label{app:dper-objective}

\noindent\hspace*{\parindent}For same-task rollout groups with $K>1$, we set all base advantages to zero when the group rewards are identical.

Using the evidence-weighted token advantage $\widehat A_{in}$ from Equation~\eqref{eq:dper-token-advantage}, we optimize the unprivileged policy with the ratio $\rho_{in}(\theta)=\pi_\theta(y_{i,n}\mid c_{i,n})/p_{0,in}$, where $p_{0,in}$ is the rollout probability defined in Equation~\eqref{eq:dper-branch-probabilities}. The objective is
\begin{equation}
\begin{aligned}
\mathcal J_{\mathrm{DPER}}(\theta)
&=\mathbb E_i\!\left[
\frac{1}{|\mathcal M_i|}\sum_{n\in\mathcal M_i}\mathcal U_{in}(\theta)
\right],\\
\mathcal U_{in}(\theta)
&=\min\!\left(
\rho_{in}(\theta)\widehat A_{in},
\operatorname{clip}(\rho_{in}(\theta),1-\epsilon,1+\epsilon)\widehat A_{in}
\right),
\end{aligned}
\label{eq:dper-policy-objective}
\end{equation}
where $\epsilon$ is the policy-ratio clipping radius. The sum and normalization include only valid assistant-token positions $\mathcal M_i$, and the expectation averages over task instances and their sampled trajectories. The rollout probabilities, privileged scores, routing decisions, and token advantages are detached and held fixed during updates on a rollout batch. They are recomputed using the next student snapshot and newly collected rollouts. Gradients update only the unprivileged policy.

\subsection{GPRM Grade Normalization}
\label{app:gprm-normalization}

\noindent\hspace*{\parindent}For reasoning step $s$, we apply softmax only to the logits of the four grade tokens $\mathcal G=\{A,B,C,D\}$ at the fixed judgment position after the generated verification rationale. The resulting probabilities $p_{s,g}$ sum to one over these four grades. These probabilities enter the expected-grade reward in Equation~\eqref{eq:gprm-trajectory-reward}, which uses weights $(1,\frac23,\frac13,0)$ for grades A through D and averages step rewards within each trajectory.

\subsection{Reward Availability and Scoring Paths}
\label{app:reward-implementation}

\noindent\hspace*{\parindent}Failure to construct the scoring input, including missing prompt data, directly returns $R_i=0$. With valid input but no evaluable process steps, GPRM is unavailable; the scorer skips GPRM, Judge, and Queue calls and returns $0.1\,\ell_{\mathrm{format},i}\ell_{\mathrm{len},i}$. A failed GPRM request or a missing or invalid score uses the same fallback and skips Judge and Queue. A valid GPRM score, including zero, instead enters normal aggregation in Equation~\eqref{eq:reward-total}, with $\beta_{\mathrm{GPRM}}$ retained in the denominator; Judge and Queue are evaluated when applicable. Normal aggregation excludes unavailable optional scores and retains valid zeros from every branch. These availability checks are distinct from process-quality judgments. The group-level hard-failure exclusion in Section~\ref{sec:interaction-rewards} still takes precedence over policy updates.

\subsection{Task-Queue Scoring}
\label{app:taskqueue-semantics}

\noindent\hspace*{\parindent}Section~\ref{sec:interaction-rewards} combines semantic agreement with queue continuity. Let $F_{\mathrm{inv}}$ reconstruct the intent expressed by a queue under task-granularity rules $\mathcal R_{\mathrm{gran}}$. In both initial and feedback-turn scoring, $\operatorname{sim}$ encodes the compared texts with Qwen3-Embedding-4B and linearly maps their cosine similarity from $[-1,1]$ to $[0,1]$.

\paragraph{Initial-queue scoring.}
\label{app:taskqueue-initial}
On the first turn, the reconstructed intent is compared with the initial request $x$:
\begin{equation}
\begin{aligned}
\widehat x_{\mathrm{queue}}^{(1)}
&=F_{\mathrm{inv}}(Q^{(1)};\mathcal R_{\mathrm{gran}}),\\
q_{\mathrm{queue}}^{(1)}
&=\operatorname{sim}(\widehat x_{\mathrm{queue}}^{(1)},x).
\end{aligned}
\label{eq:taskqueue-first-turn-reward}
\end{equation}
This comparison evaluates intent coverage without requiring an exact match to a reference queue. For queues containing coarse-grained action labels, the score reflects task-type coverage rather than detailed argument correctness or task completion. No continuity score is used on the first turn.

\paragraph{Availability and missing outputs.}
Initial and feedback-turn scores use Equations~\eqref{eq:taskqueue-first-turn-reward} and~\eqref{eq:taskqueue-combined-reward}, respectively, with weights from Section~\ref{sec:experimental-setup}. Available components are reweighted to sum to one, retaining evaluated zeros. A turn has no Queue score if all components are unavailable or the output lacks the task-queue field; the format factor separately assesses required fields. These rules follow the early-exit checks in Appendix~\ref{app:reward-implementation}.

\paragraph{Queue inverse rewriting.}
The inverse rewrite preserves the listed tasks, any arguments present, batch structure, and execution-order or parallelism constraints, without adding, omitting, or correcting tasks. It uses only the queue itself. We use JoyAI-VL for this rewriting in both initial and feedback-turn scoring.

\begin{table}[htbp]
\centering
\caption{Illustrative TaskQueue inverse-rewriting prompt for JoyAI-VL. The placeholder \texttt{\{task\_queue\}} is replaced with the current queue.}
\label{tab:taskqueue-inverse-prompt}
\begin{tcolorbox}[
  colback=black!2,
  colframe=black!40,
  colbacktitle=black!8,
  coltitle=black,
  boxrule=0.4pt,
  arc=1mm,
  left=8pt,right=8pt,top=6pt,bottom=6pt,
  fonttitle=\small\bfseries,
  fontupper=\small\ttfamily,
  title={TaskQueue inverse rewriting}
]
\raggedright
Rewrite the following TaskQueue as a concise natural-language description of the tasks it specifies. Preserve all tasks, arguments, batch structure, and execution-order or parallelism constraints. Do not add, omit, or correct tasks. Return only the description.

\medskip
TaskQueue: \{task\_queue\}
\end{tcolorbox}
\end{table}
\FloatBarrier

For a turn with a nonempty observation block and a preceding queue, let $o^{(1:r)}=(o^{(1)},\ldots,o^{(r)})$ denote the full observation history available before the current queue update, including user messages, visual observations, execution feedback, and errors. Given the initial request $x$ and this history, $F_{\mathrm{obs}}$ rewrites the task requirements to reflect the remaining tasks after accounting for completed work, user revisions, and cancellations. The semantic score compares the updated intent with the intent reconstructed from the current queue:
\begin{equation}
\begin{aligned}
\widehat x_{\mathrm{obs}}^{(r)}&=F_{\mathrm{obs}}(x;o^{(1:r)}),\\
\widehat x_{\mathrm{queue}}^{(r)}&=F_{\mathrm{inv}}(Q^{(r)};\mathcal R_{\mathrm{gran}}),\\
q_{\mathrm{obs}}^{(r)}&=\operatorname{sim}(\widehat x_{\mathrm{obs}}^{(r)},\widehat x_{\mathrm{queue}}^{(r)}).
\end{aligned}
\label{eq:taskqueue-observation-reward}
\end{equation}
To measure continuity, let $E^{(r-1)}$ be the verified execution outcome for the previous head batch $B_1^{(r-1)}$, with status $\nu(E^{(r-1)})$ equal to complete, incomplete, or partial. Incomplete includes failed or unexecuted heads; $\nu$ is determined from execution feedback rather than the queue proposed by the policy. Let $\bigl[B_{\mathrm{rem}}^{(r-1)}\bigr]$ contain the nonempty unfinished part of the previous head after partial execution, as determined by $E^{(r-1)}$, preserving the original relative order of the remaining tasks. With $\rVert$ denoting queue concatenation, define the execution-conditioned reference queue as
\begin{equation}
Q_{\mathrm{ref}}^{(r)}=
\begin{cases}
[], & Q^{(r-1)}=[],\\
\operatorname{Tail}(Q^{(r-1)}), & \nu(E^{(r-1)})=\mathrm{complete},\\
Q^{(r-1)}, & \nu(E^{(r-1)})=\mathrm{incomplete},\\
\bigl[B_{\mathrm{rem}}^{(r-1)}\bigr]\rVert\operatorname{Tail}(Q^{(r-1)}),
& \nu(E^{(r-1)})=\mathrm{partial}.
\end{cases}
\label{eq:taskqueue-reference-queue}
\end{equation}
Let $M_r$ count batch pairs that match exactly at corresponding positions in $Q_{\mathrm{ref}}^{(r)}$ and $Q^{(r)}$. A pair matches only when both batches contain the same number of tasks and identical task entries in the same order. Matching permits no reordering of batches or their tasks, including parallel tasks. In the coarse-grained representation, entries are action labels rather than tool-call arguments. Semantic similarity is evaluated separately. When at least one queue is nonempty, the continuation score is
\begin{equation}
q_{\mathrm{cont}}^{(r)}
=\frac{M_r}{\max\{|Q_{\mathrm{ref}}^{(r)}|,|Q^{(r)}|\}},
\label{eq:taskqueue-continuation-reward}
\end{equation}
where queue length counts task batches. The score is defined as 1 when both queues are empty. Equation~\eqref{eq:taskqueue-combined-reward} combines this score with semantic agreement on feedback turns.

\paragraph{Cross-turn aggregation.}
The trajectory-level TaskQueue reward averages all evaluated turn scores, including zeros. For a single-turn trajectory, it equals the initial score when evaluable. If no turn is evaluable, normal content aggregation omits the Queue component and renormalizes the remaining weights.

\clearpage
\section{Product Prompt and Interaction Templates}
\label{app:developer-prompt}

\noindent\hspace*{\parindent}Table~\ref{tab:product-developer-prompt} presents an English structural template of the product developer prompt. The prompt specifies five elements: role definition, input specification, output requirements, capability registry, and execution policies. In the notation of Section~\ref{sec:data-construction}, the role and communication interface instantiate $I_P$, the capability registry instantiates $S_P$, and the execution policies encode $W_P$ and $C_P$. Placeholders (enclosed in curly braces) represent configuration-specific content, including the instruction source and output recipient; concrete tool definitions and policy instances are omitted.

\begin{table}[htbp]
\centering
\caption{Product developer prompt template with configurable role, interaction interface, capabilities, and execution policies.}
\label{tab:product-developer-prompt}
\begin{tcolorbox}[
  colback=black!2,
  colframe=black!40,
  colbacktitle=black!8,
  coltitle=black,
  boxrule=0.4pt,
  arc=1mm,
  left=8pt,right=8pt,top=6pt,bottom=6pt,
  fonttitle=\small\bfseries,
  fontupper=\small\ttfamily,
  title={Developer prompt structure}
]
\raggedright
\textbf{1. Role Definition}

Role and embodiment: \{role\_definition\}.

Execute assigned tasks using the registered capabilities and prescribed execution policies.

\medskip
\textbf{2. Input Specification}

Instruction source: \{instruction\_source\}.

Task representation: \{task\_input\_schema\}.

\medskip
\textbf{3. Output Requirements}

Recipient: \{output\_recipient\}.

Style: objective, concise, and factual.

Execution report format: \{execution\_report\_schema\}.

\medskip
\textbf{4. Capability Registry}

For each registered capability, specify:

Name: \{capability\_name\}.

Description: \{capability\_description\}.

Parameters: \{parameter\_schema\}, including types, units, and default values where applicable.

\medskip
\textbf{5. Execution Policies}

For each applicable policy, specify:

Policy name: \{policy\_name\}.

Execution procedure and conditions: \{execution\_rules\}.

Operational constraints: \{applicable\_constraints\}.
\end{tcolorbox}
\end{table}
\FloatBarrier

\clearpage
\subsection{Chat-Template Serialization}
\label{app:chat-template}

\noindent\hspace*{\parindent}The developer prompt is the content of a distinct \texttt{developer} message. Table~\ref{tab:developer-chat-template} reproduces the corresponding Jinja branch and illustrates the resulting input sequence. The role name follows \texttt{<|im\_start|>}, and the message ends with \texttt{<|im\_end|>}; the product specification is supplied as text within these boundaries. This serialization distinguishes the shared system instructions from the product specification without changing the message delimiters. Tool results are serialized inside \texttt{<tool\_response>} blocks within user messages, preserving their source through these tags.

\begin{table}[htbp]
\centering
\caption{Developer-role serialization. Top: the exact branch from the chat template. Bottom: a representative input with \texttt{add\_generation\_prompt} enabled and thinking enabled. Braces denote substituted content; optional tool definitions, visual-token expansion, and interaction history are omitted.}
\label{tab:developer-chat-template}
\begin{tcolorbox}[
  colback=black!2,
  colframe=black!40,
  colbacktitle=black!8,
  coltitle=black,
  boxrule=0.4pt,
  arc=1mm,
  left=8pt,right=8pt,top=6pt,bottom=6pt,
  fonttitle=\small\bfseries,
  fontupper=\footnotesize,
  title={Developer role in the chat template}
]
\textbf{Jinja branch}
\begin{verbatim}
{%- elif message.role == "developer" %}
    {%- set dev_content = render_content(message.content, true) %}
    {{- '<|im_start|>developer\n' + dev_content + '<|im_end|>' + '\n' }}
\end{verbatim}
\textbf{Serialized model input}
\begin{verbatim}
<|im_start|>system
{shared_interaction_protocol}<|im_end|>
<|im_start|>developer
{product_specification}<|im_end|>
<|im_start|>user
{task_request_and_multimodal_content}<|im_end|>
<|im_start|>assistant
<think>
\end{verbatim}
\end{tcolorbox}
\end{table}
\FloatBarrier

\subsection{ReAct Assistant Output Template}
\label{app:joint-assistant-output}

\noindent\hspace*{\parindent}Table~\ref{tab:joint-assistant-output} illustrates the ReAct decisions described in Section~\ref{sec:product-sft}. A single assistant message combines the shared prefix with either continuation. The \texttt{task} block is included in either branch when required by the interaction protocol and omitted otherwise. Tool execution feedback conditions the next policy decision; direct responses address the user without issuing a tool call.

\begin{table}[htbp]
\centering
\caption{Illustrative ReAct assistant output. The shared prefix is followed by one of the two continuations. Braced entries are placeholders; the \texttt{task} block is included when required by the interaction protocol.}
\label{tab:joint-assistant-output}
\begin{tcolorbox}[
  colback=black!2,
  colframe=black!40,
  colbacktitle=black!8,
  coltitle=black,
  boxrule=0.4pt,
  arc=1mm,
  left=8pt,right=8pt,top=6pt,bottom=6pt,
  fonttitle=\small\bfseries,
  fontupper=\small,
  title={ReAct assistant output: tool call or direct response}
]
\textbf{Shared prefix}
\begin{verbatim}
<think>{reasoning based on the current context}</think>
<task>
[["{current_action_label}"], ["{pending_action_label}"]]
</task>
\end{verbatim}
\textbf{Continuation A: tool call}
\begin{verbatim}
<tool_call>
{"name": "{tool_name}",
 "arguments": {"{argument_name}": "{argument_value}"}}
</tool_call>
\end{verbatim}
\textbf{Continuation B: direct response}
\begin{verbatim}
{answer, clarification, or refusal addressed to the user}
\end{verbatim}
\end{tcolorbox}
\end{table}
\FloatBarrier

\end{document}

%% file: authors.tex
\author[*]{Gangyi Li}
\author[*]{LuYao Wang}
\author[*]{Jingfu Sun}
\author[*]{Jingyi Yuan}
\author{Hang Liu}
\author{Hao Zong}
\author{Yujia Hu}
\author[*,\dagger]{Tingting Gao}
\author[\dagger,\ddagger]{Jianyong Wang}
\author[\dagger,\ddagger]{Xu Yang}

\affiliation{JD.COM}
\contribution[*]{Equal Contribution}
\contribution[\dagger]{Project Lead}
\contribution[\ddagger]{Corresponding Authors}


%% file: abstract.tex
Embodied agents must translate multimodal observations and user requests into executable actions while adapting to product constraints, changing goals, and execution feedback. We present JoyInside-VLM-Agent, a ReAct-style multimodal agent developed through product-conditioned trajectory construction and staged policy learning for robotic platforms and smart hardware. A closed-loop sandbox generates multimodal interaction trajectories covering goal revisions and failure recovery, which are checked against execution outcomes and semantic criteria. Supervised fine-tuning and multi-teacher on-policy distillation precede reinforcement learning with DPER (Dual-Source Privileged On-Policy Self-Distillation with Token-Level Evidence Routing). DPER routes token-level evidence between a frozen privileged teacher and a privileged student sharing the rollout policy's parameters, modulating reward-derived token advantages while preserving their signs. Training rewards combine generative process evaluation, multi-judge rankings, and task-queue consistency. JoyInside-VLM-Agent achieves a BFCL v4 overall score of 77.44, exceeding the evaluated baselines; results on the other public benchmarks are mixed. On JoyInside-Devices and JoyInside-Robotics, the evaluated system, which includes domain-specific skill refinement, leads the baselines on all reported metrics. Two real-world quadruped case studies illustrate visually grounded search, replanning after execution failure, and explicit failure reporting.